# *Detecting Shortcut Learning for Fair Medical AI using Shortcut Testing*

Alexander Brown[1,*], Nenad Tomasev[2], Jan Freyberg[3], Yuan Liu[3], Alan Karthikesalingam[3], Jessica Schrouff[3,#]

[1]UCL Institute of Child Health
*Work performed while at Google Research
[2]DeepMind
[3]Google Research
[#]Now at DeepMind

**Corresponding author**: Jessica Schrouff, schrouff@google.com

# Abstract

Machine learning (ML) holds great promise for improving healthcare, but it is critical to ensure that its use will not propagate or amplify health disparities. An important step is to characterize the (un)fairness of ML models - their tendency to perform differently across subgroups of the population - and to understand its underlying mechanisms. One potential driver of algorithmic unfairness, shortcut learning, arises when ML models base predictions on improper correlations in the training data. Diagnosing this phenomenon is difficult as sensitive attributes may be causally linked with disease. Using multi-task learning, we propose a method to directly test for the presence of shortcut learning in clinical ML systems, and demonstrate its application to clinical tasks in radiology and dermatology. Finally, our approach reveals instances when shortcutting is not responsible for unfairness, highlighting the need for a holistic approach to fairness mitigation in medical AI.

# Introduction

Machine learning (ML) promises to be a powerful approach in many healthcare settings, with models being designed for a variety of diagnostic and prognostic tasks. A risk of harm from machine learning models is unfairness, as variation in model behavior for patients with different sensitive attributes (Figure 1a) has the potential to perpetuate or amplify existing health inequities [1,2]. This has been observed in multiple clinical settings [3–5] and remains a major topic of research. While the definition of what constitutes fairness may vary widely across fields [6], here we focus on the expression of fairness as equal model performance across patient subgroups defined by sensitive attributes [7].

On the other hand, machine learning models may utilize information about sensitive attributes (such as age, sex, or race), to improve model performance [8] in ways that may be justifiable where attributes correlate with disease risk/presentation in the deployment population. For instance, androgenetic alopecia is more prevalent in men and breast cancer more common in women; keloid scarring is more common in skin of color[9] and melanoma more common in lighter skin tones. In such settings, ignoring or ablating attribute information may decrease clinical performance.

However, the use of information about sensitive attributes can also be harmful - in particular due to the phenomenon of shortcut learning[10]. This refers to ML models relying on spurious associations in training datasets to learn prediction rules which generalize poorly, particularly to new populations or new settings. Shortcut-based decision rules are also likely to amplify errors in atypical examples, such as male patients with breast cancer, or melanoma in dark-skinned individuals. While shortcut learning is typically evaluated by focusing on the performance of a model in different populations or environments[10], shortcuts based upon sensitive attributes have the risk to exacerbate model unfairness, and to further health disparities. Our work hence investigates how shortcuts might affect model fairness, in addition to performance.

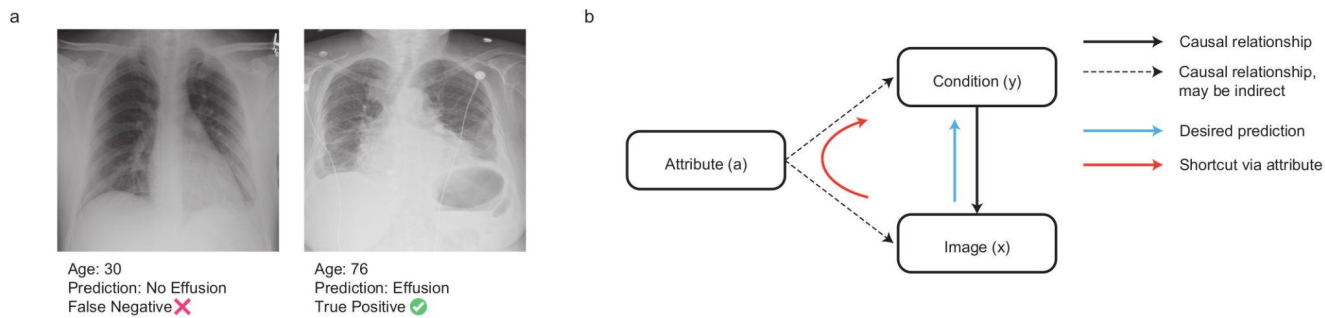

Figure 1: Is a model unfair due to shortcutting? (a) Examples of correct and incorrect predictions that may be influenced by age shortcut learning, in a chest x-ray application detecting Effusion. In this example, the prediction is incorrect for a patient with atypical age, raising the possibility of shortcut learning. (b) Simplified diagram illustrating how shortcutting may occur. In this schematic, the presence or absence of a particular condition *y* will produce changes in the image *x*; we therefore wish to train a model that can predict *y*, given *x* (blue arrow). However, an attribute *a*, such as age, may alter both the risk of developing a given condition, as well as the image. This need not be a directly direct relationship (dotted arrows). A model may learn to predict the presence of a condition by using the attribute (red arrow). When these correlations are not considerably beneficial for model performance, we consider that this path is (at least partly) a shortcut. Source data are provided as a Source Data file.

Concerns about ML models exploiting shortcuts based on sensitive attributes have been amplified by the observation that ML models can predict these attributes from clinical data, without the need for attributes to be directly inputted to the model. For example, models can be trained to predict sex or age from medical images [11,12], and may even encode information about sensitive attributes when this was not the objective of ML training [13]. However, the fact that a ML model encodes information about sensitive attributes does not necessarily mean that it uses this information to make clinical predictions [14], or that such use results in shortcut learning.

Previous work on shortcut learning[10] has focussed on sensitive attributes that are likely to encode spurious correlations. In this context, any reliance by the model on the sensitive attribute may be considered to be a shortcut. However, this approach does not generalize to cases where a sensitive attribute may be, at least partly, causally related to the outcome. In this work, we posit that the effect of sensitive attributes on the model is the sum of biological, potentially causal effects which might improve model performance, and shortcut learning, which might be harmful. In this context, we redefine shortcut learning to be an effect of the sensitive attribute that does not considerably improve performance (as defined by the user) but affects fairness. By intervening on the degree to which a model can encode a sensitive attribute, we demonstrate a method that assesses whether such encoding indicates the presence of shortcut learning, appropriate use of sensitive attributes, or is artefactual.

The main contribution of this study is an approach that represents a practically applicable framework for studying and mitigating shortcut learning in clinical ML models. This addresses an unmet need among practitioners when trying to develop fair and safe clinical AI. To illustrate our method, we refer to radiology and dermatology applications and focus on age as a sensitive attribute, since aging is linked to disease risk across a wide variety of medical conditions, making it harder to reliably establish whether a model is relying on shortcuts. In addition, we follow up on prior work by Gichoya et al.[13] investigating the effect of race encoding to understand how this encoding affects model performance and fairness. In this case, we consider race to be a social construct, akin to a spurious correlation[13].

# Results

### Prediction models encode age and are unfair

Using an open-source chest x-ray (CXR) dataset, we trained separate deep learning models for each of three CXR prediction targets (Atelectasis, Effusion, Abnormal, Supplementary Figure 1a). The model architecture comprised a feature extractor followed by a clinical prediction head[8]. Using a transfer learning paradigm (i.e. freezing the weights of the feature extractor and training the model to predict age), we then characterized the amount of age information contained in the penultimate layer of each model (see Methods).

We find that transfer models were able to predict age (Supplementary Figure 1b; Effusion 11.9 ± 0.47 years; Atelectasis 11.3 ± 0.28; Abnormal 11.4 ± 0.44, age MAE on held-out test set) significantly better than chance (permutation test, p<0.0001 for all models). We estimated experimental bounds on these results by training the full architecture to predict age (lower error bound) and to predict the mean age across training samples (upper error bound, see Methods). Models performed better than the upper error bound of 13.6 years, but worse than the lower error bound defined by a direct prediction model of 6.4 ± 0.23 years.

We then estimated algorithmic fairness as defined by separation[7], i.e. discrepancies in model's true positive and false positive rates based on age. All tasks produced models with a bias in performance according to age based on separation (Supplementary Figure 1c). The observed separation values, in the range of 0.01-0.02, correspond to around an 11-22% performance difference per decade of life, a discrepancy that we feel is likely to be unacceptable to users in the absence of other considerations.

Our findings demonstrate that CXR models do learn to encode age, despite not being trained to do so. In addition, the performance of the models varies systematically with age, exhibiting unfairness. However, it is not possible to infer from these observations alone that the encoding of age is a driver of age-related unfair performance - which would be required for shortcut learning.

### Intervening on attribute encoding affects fairness metrics

In order to test the degree to which such encoding may drive unfairness via shortcut learning, we performed an intervention that varies the amount of age encoding in the feature extractor and assessed the effect of this intervention on model fairness. We refer to this analysis as Shortcut Testing or ShorT. Multiple techniques can be used to perform this intervention (e.g. group-DRO[15], data sampling or reweighting). Based on prior work on adversarial learning[10,16–19], we selected to vary the scaling of the gradient updates from the age head in a multitask learning paradigm (see Methods). We focus here on results for the Effusion label (Figure 2a); similar results were obtained for Atelectasis and Abnormal (Supplementary Figure 2).

We were able to vary the amount of age information encoded in the feature representation across a wide range of MAE values, covering the region between the upper and lower bounds on age prediction error (Figure 2b). By plotting the fairness of the resulting models against performance, we compared the impact of altered age encoding on these critical properties (Figure 2c). The ideal model would have high performance and low separation, in the top left corner of the scatter plot.

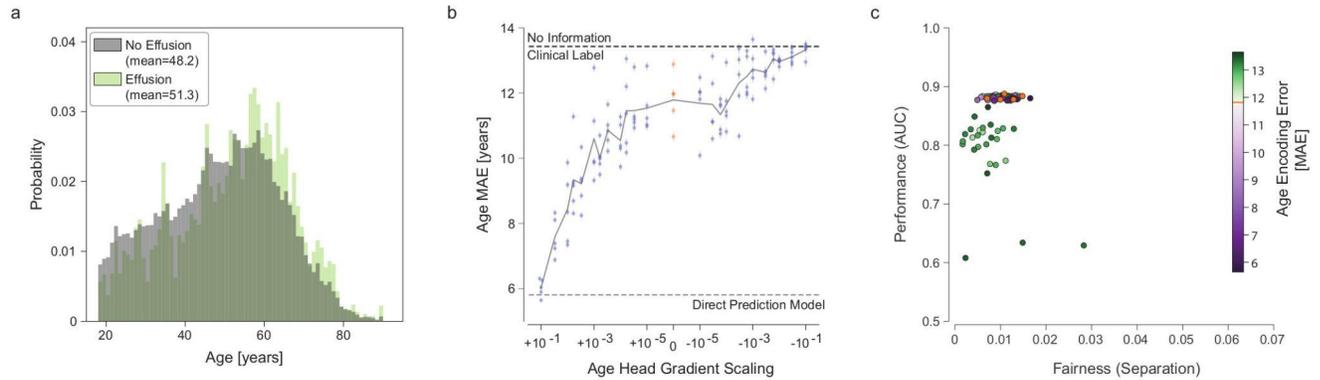

Figure 2: Intervening on age encoding using multitask learning. (a) The distribution of ages for positive (light green) and negative (gray) examples for Effusion in the training set. (b) The effect of altering the gradient scaling of the age prediction head on age encoding (as determined by subsequent transfer learning). For large positive values of gradient scaling (left), the models encoded age strongly, with a low MAE that approached the performance of a dedicated age prediction model (empirical LEB). For large negative values of gradient scaling (right), the age prediction performance of the multitask model approached the empirical UEB. Baseline models (with zero gradient scaling from the age head, equivalent to a single task condition prediction model) are shown in orange. Each dot represents a model trained (25 values of gradient scaling times 5 replicates), with error bars denoting 95% confidence intervals from bootstrapping examples (n=17,723) within a model. (c) Fairness and performance of all replicates (n=125). The degree of age encoding by the particular replicate is color-coded, with purple dots denoting more age information, and green dots less information, than the baseline model without gradient scaling (in orange, overlapping with the purple dots in this case).

In this case, we found that increasing age encoding relative to baseline (purple dots in Figure 2c) does not noticeably affect model performance. Reducing the age encoding (green dots) appeared to slightly improve fairness properties, but at the cost of reducing overall model performance. We quantitatively analyze this effect below.

### ShorT efficiently detects shortcut learning

By definition, shortcut learning is driven by correlations between an attribute such as age, and condition prevalence and appearance. Whilst not identical, the distribution of ages of patients with and without effusion in the NIH dataset is quite similar, with a mean age of 48.2 for patients without effusion, and 51.3 for patients with effusion (Figure 2a).

We therefore applied the ShorT method to datasets in which this correlation was strengthened or weakened, simulating the results expected in contexts in which shortcut learning is respectively more or less likely. This was achieved by simple subsampling of the training data to create a biased dataset, and a balanced dataset (see Methods). We expect that a greater correlation (biased dataset) will lead to a strong pattern of shortcut learning while a weaker correlation (balanced dataset) should not lead to shortcut learning.

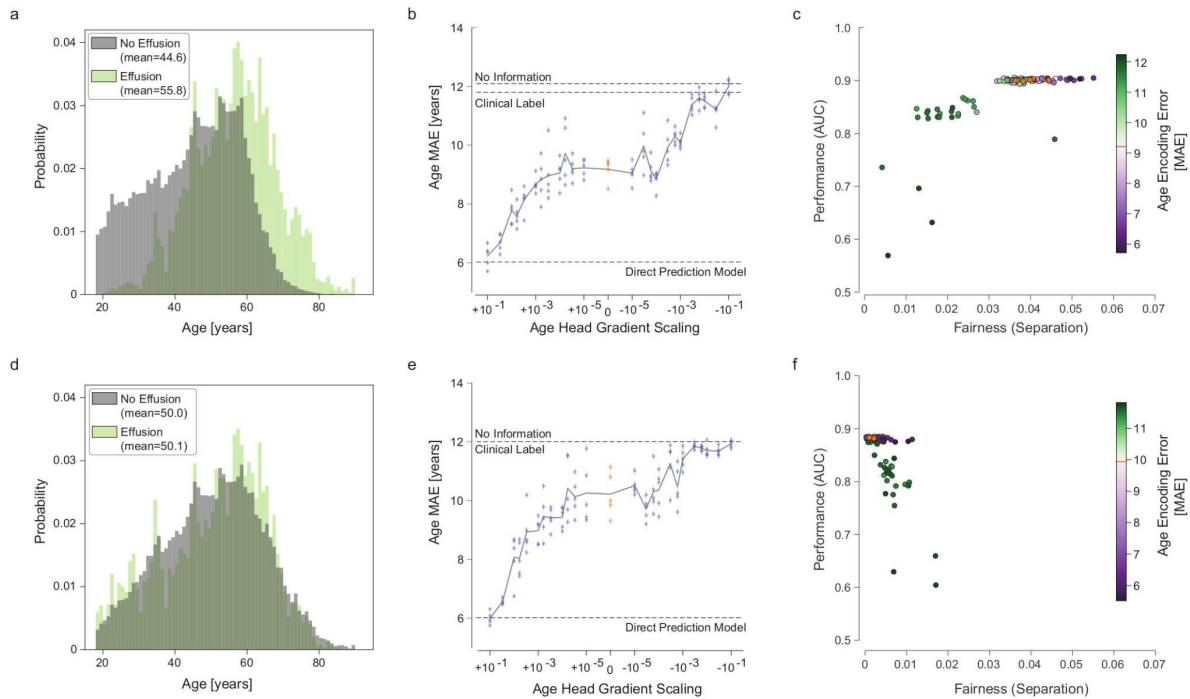

Figure 3: Effect of dataset perturbation. Results are presented in the same manner as for figure 2, but for a subsampled dataset inducing a larger age disparity between classes (biased dataset, a-c, n=15,148), and a balanced dataset (d-f, n=16,093).

In the biased dataset, the correlation between age and the effusion label was artificially strengthened by preferentially dropping examples of older patients without effusion, or younger patients with effusion. We sought to induce an approximate decade of difference between classes; owing to the inherent stochasticity of this method, the actual gap was 11.2 years (Figure 3a). When trained on this perturbed dataset, baseline models without additional age heads represented age more strongly (9.18 years age MAE for models trained on biased dataset, vs 11.8 years for models trained on full dataset); however altering the gradient scaling of the age head in multitask models still resulted in a wide range of age encoding strengths (Figure 3b). Clinical task performance was similar, albeit slightly higher in the biased dataset (mean AUC 0.901 vs 0.882 in the original dataset). This is to be expected, since the separation by attribute creates further information that can be used to make more accurate predictions. However, the fairness of the models was degraded strikingly (Figure 3c). This disparity could be obviated to some degree by gradient reversal for age - with separation approximately halved in models with poorer age representation, with only a slight decrease in overall model performance (green cluster).

On the other hand, removing age differences due to prevalence (the balanced dataset, Figure 3d-f) resulted in models that performed at a similar level to the baseline model (mean AUC 0.883 in the balanced dataset, 0.882 original) and that these were fairer than models trained on the original dataset.

We found similar results on Atelectasis, and Abnormal labels when using biased datasets (Supplementary Figure 3), with a clear pattern of dependence of model fairness relating to age encoding. Our results are also replicated for effusion with other fairness metrics (Supplementary Figure 4).

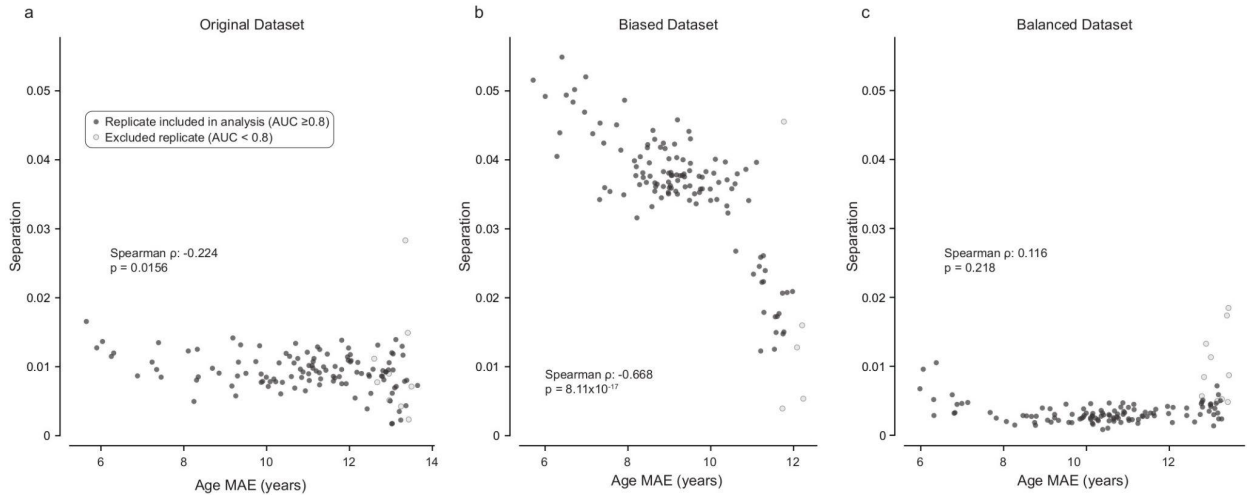

Figure 4: Quantifying the systematic effect of age encoding on fairness. In each panel, replicates are shown as individual points (n=125), with age mean absolute error (MAE) on the x-axis, and separation on the y-axis. Note that increasing values on the x-axis correspond to a reduction in age encoding; increasing values on the y-axis denote worsening fairness. Therefore, a correlation between encoding and unfairness would be observed as a negative correlation. (a) Original dataset. For this dataset, models which encoded age to a lesser degree (right) tended to be slightly more fair (lower separation), as reflected by a statistically significant spearman rank correlation (two-sided). (b) The same analysis for the biased dataset. In this case, the correlation between age encoding and unfairness is markedly strengthened. (c) Balanced dataset. There was no significant correlation between age encoding and unfairness in this dataset. In all analyses, replicates with unacceptable performance (AUC < 0.8, arbitrary threshold) were excluded (open dots).

In order to quantify the statistical dependence of unfairness on age encoding - a signature of shortcut learning - we calculated the Spearman rank correlation between these two variables (Figure 4). For the original dataset (Figure 4a), there was a small but statistically significant correlation between encoding and unfairness ($\rho$=-0.224, p=0.0156), indicating the presence of shortcut learning. This was amplified in models trained on the biased dataset (Figure 4b), indicating the presence of significantly stronger shortcutting when training with this biased dataset ($\rho$=-0.668, p=8.11e-17). Conversely, the correlation coefficient in the balanced dataset was not significant (Figure 4c, $\rho$=0.116, p=0.218), indicating no systematic impact of age encoding on fairness in models trained on this dataset (for details, see Supplementary Methods). Based on these results, our approach seems to efficiently detect shortcut learning.

## Shortcut learning cannot be identified by attribute encoding alone

On the other hand, we found that the amount of age information encoded in the model bears little relation to the fairness of the model when comparing datasets (Figure 4, compared directly in Supplementary Figure 5). For an age MAE of 8.5-9.5 years, models trained on a balanced dataset were almost perfectly fair, with an average separation coefficient of 0.0016 (range 0.0003-0.0035, n=24), corresponding to an average 1.6% disparity in performance over a decade of life. In contrast, models trained on a biased dataset had a mean separation coefficient of 0.0384, (range 0.0335-0.0461, n=43), corresponding to an average 47% disparity over a decade of life. Thus, in this case, it is clear that the performance of an attribute transfer model alone is insufficient to make any predictions regarding the fairness of the model. Rather, testing directly for the effect of encoding on fairness, reveals the presence of shortcut learning.

## ShorT detects shortcutting by race in cardiomegaly predictions

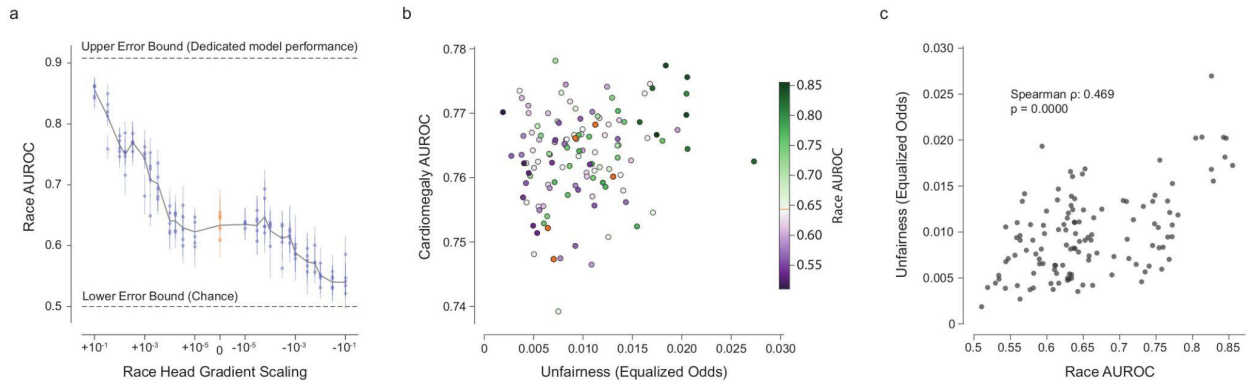

Figure 5: Unfair model performance resulting from shortcut learning in a cardiomegaly classifier. (a) The effect of altering the gradient scaling of the binary race prediction head on race encoding (as determined by the race prediction AUROC of subsequent transfer learning). Each dot represents a model trained (25 values of gradient scaling times 5 replicates), with error bars denoting 95% confidence intervals from bootstrapping examples within a model (n=3818 independent patients). (b) AUC vs fairness (equalized odds) plot for Cardiomegaly. There exist models that are fairer and as performant as the baseline model (orange dots). (c) ShorT analysis demonstrates that unfairness is significantly correlated with race encoding in this example (two-sided Spearman correlation, n=123 technical models).

Using the public dataset CheXpert[20], we predicted cardiomegaly from chest x-rays as a binary outcome. Following the work in Gichoya et al.[13], we investigated whether the representation of race (self-reported binary attribute, Black or White) led to shortcutting in our model. Our results showed that shortcutting was present ($\rho$=0.469, p<0.001, Figure 5), with fairness (here estimated via equalized odds) depending strongly on the model's encoding of race (estimated by the AUROC of race prediction). Contrary to the biased dataset for Effusion, we however note that there is no apparent trade-off between the model's clinical performance and fairness. In this case, ShorT provides fairer but as performant alternatives to the original model in addition to the detection of shortcutting (purple dots in top left corner of Figure 5b).

## Beyond shortcut learning: Acne prediction in a Dermatology model

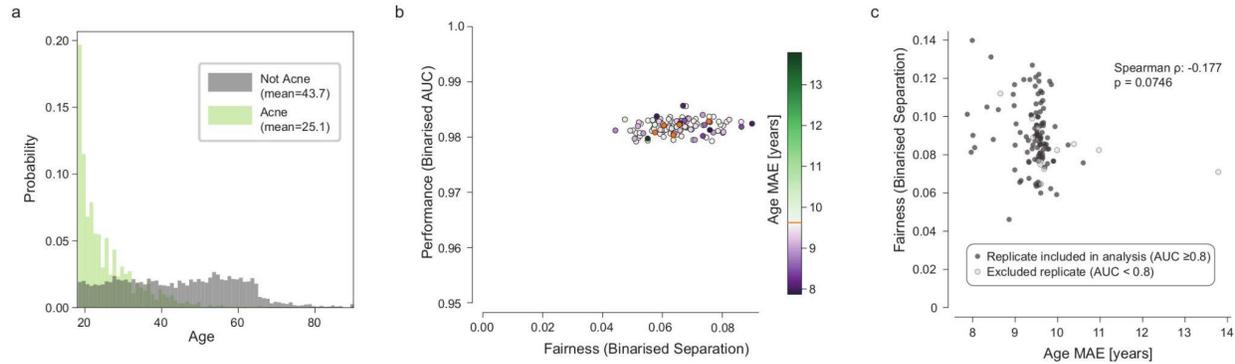

Figure 6: Unfair model performance not resulting from shortcut learning in a dermatology classifier - despite strong attribute-condition correlation. (a) Age distribution for examples with Acne (light green) and all other conditions (gray) in the training set. Note the significantly lower mean age for patients with Acne, as would be expected clinically. (b) AUC vs separation plot for Acne. AUC is binarised by using the score for Acne vs all; separation is calculated on a binarised prediction (top1). (c) ShorT analysis does not demonstrate that separation is significantly correlated with age encoding in this example (two-sided Spearman correlation).

Lastly, we applied our approach to a multiclass prediction model in dermatology (see Supplementary Methods) similar to that published in [8], examining a single class using a binarised form of our analysis. The most common condition in this dataset is Acne, which is strongly correlated with age (18.6 year difference in mean age between patients with Acne vs other conditions; Figure 6a). The multiclass model represents age strongly, with a mean age MAE of 9.58 years across 5 baseline replicates, compared to an experimental LEB of 7.32, and experimental UEB of 13.29 years (Supplementary Figure 7).

Baseline models with no gradient updates from the age head show clear discrepancies in predicting the presence or absence of Acne (separation range 0.0576-0.0755, Figure 6b, orange dots). This corresponds to a separation differential of up to 53% per decade.

However, despite the bias in the training set, strong age encoding in the model, and unfair performance, we found that varying the amount of age encoding did not result in a systematic change in fairness properties in this case (Figure 6c, $\rho$=-0.177, p=0.0746). Although the models are considerably less sensitive for Acne for older patients, the specific cause for this does not appear to be shortcut learning. There are a variety of other mechanisms which can lead to unfair performance, discussed below. However, multitask learning may still prove valuable in identifying models with high performance and better fairness properties.

# Discussion

Shortcut learning poses significant challenges for machine learning in healthcare, where predictions based on spurious correlations pose significant concerns regarding safety[21,22] and fairness [3,4,12]. However, identifying whether shortcut learning is responsible for model unfairness is challenging, especially when sensitive attributes such as age may be causally linked to the clinical task.

In this paper, we propose a practical method to directly test for the presence of shortcut learning during routine fairness assessment of clinical AI systems when attributes might be causally related to the outcome. We show that the degree to which models encode for a sensitive attribute - which has previously been suggested as an analytical approach[13,23] - is a poor measure of the degree to which that attribute may be used. This supports previous work demonstrating that the presence of sensitive attribute encoding by a model may arise incidentally[14]. Rather, we focus not on the encoding itself, but on the degree to which the fairness of a model's predictions depend on such encoding.

While our approach is primarily designed to investigate shortcut learning, a useful by-product is that it creates a family of models mitigated with varying degrees of gradient reversal. However, the particular choice of mitigation strategy will depend on the dataset and task, and more complex strategies may prove more effective or be more feasible[24]. We also demonstrate that our method can indicate when shortcut learning is unlikely responsible for model unfairness, which should prompt the exploration of alternative mitigation strategies.

Prior to applying our method, it is critical to select an appropriate fairness criterion. Fairness criteria are commonly classified into independence, separation and sufficiency [7]; the choice of metric will depend upon the particular clinical task, as well as wider societal context [25]. In this study we selected separation since age may be causally linked to many of the CXR findings our models were trained to predict, making independence an inappropriate choice. Selection of an appropriate metric also requires a deep understanding of how bias and inequity may be present in clinical environments and datasets. We therefore recommend consultation with subject experts, patient groups, and literature review; to identify plausible links between sensitive attributes and the clinical prediction target. This is likely to be improved by participatory problem formulation [26], but remains a challenging and open problem in the field to address comprehensively. We also note that multiple fairness metrics can be estimated and their correlation with the encoding of the sensitive attribute compared. This might provide further information on which mitigation might help, or select a multi-dimensional trade-off[27]. Although demonstrated in the context of separation, our framework is equally applicable to other fairness metrics.

Next, we would advocate for an initial exploratory analysis of the data, focusing on correlations that may be present between sensitive attributes and the prediction class. Such correlations are likely to drive shortcut learning. Intuitively, the greater the correlation between sensitive attribute and prediction target, the more likely it is that shortcut learning will occur. However, this is likely to be true only for similar datasets within a single task. For prediction of Effusion on CXR, there is a clear pattern of increasing shortcut learning as the age-effusion correlation increases in subsampled datasets (Figure 4, Supplementary Figure 5). However, shortcut learning is still observed in the original dataset, in which there is only a small difference in age between the positive and negative classes. In the case of Acne prediction in a dermatological model, no shortcut learning was observed despite a far greater correlation between age and the condition. Therefore, correlation analysis alone is insufficient to detect shortcut learning.

Another approach that has been advocated is to investigate the degree to which models encode the sensitive attribute. As has been shown for multiple sensitive attributes [13,14], we demonstrated that clinical AI models were able to encode information about age or race, despite not being explicitly trained to do so. Therefore, providing the model with information about the age or race of a patient (as an auxiliary input) is not required for shortcut learning to occur.

Such encoding is often assumed to demonstrate that the model has learnt to represent attributes so as to use them as shortcuts for predictions. However, whilst the presence of encoding is necessary for shortcut learning to occur, it does not provide conclusive evidence that models are basing diagnostic decisions on the encoded information using shortcut learning [14]. Our results demonstrate that the degree to which models encode sensitive information was not predictive of the fairness of the models, in either the CXR or dermatology tasks. To our knowledge, this has not previously been empirically demonstrated in medical AI.

We therefore developed a method to directly test whether unfair performance is driven by the observed encoding of the sensitive attribute. We demonstrate this by varying the strength of this encoding using an additional demographic prediction head with variable gradient scaling. Models whose fairness did rely on the sensitive attribute will be systematically affected by changes in the degree of sensitive attribute encoding; models in which the encoding is incidental should not be affected in the same manner. Our approach does not quantify the strength or impact of such shortcut learning, but merely whether it is statistically present in the training setup. Another important note is that ShorT relies on an intervention that varies the encoding of the sensitive attribute in the feature extractor (here gradient scaling). If the intervention does not cover a wide range of age encoding levels (between lower and upper error bounds), the correlation would not be reliable. On simulation data (see Supplementary Note 1), we observed that there could be a high variance in the model's encoding of the sensitive attribute based on random factors such as the random seed[28] when the sensitive attribute and the outcome had similar signal-to-noise ratio. Therefore, verifying that the intervention consistently modifies the encoding of the sensitive attribute is needed before estimating its relationship with model fairness. Similarly, ShorT depends on the evaluation of a selected fairness metric. If this evaluation is under-powered (e.g. some subgroups being small) or highly variable, ShorT might provide misleading results. This concern is broadly applicable to any fairness evaluation. Future work could consider how to incorporate variance or confidence intervals of fairness evaluations in the formulation of ShorT.

Rather than imposing a novel or particular architecture, our approach involves the addition of a demographic prediction head to the model under investigation, in order to generate a family of similar models with altered reliance upon attribute encoding. This family of models is then used primarily to define whether shortcut learning is occurring; the multihead models are not necessarily intended to be used instead of the base model.

We note that even if shortcut learning is not detected, encoding may present intrinsic ethical concerns and potential for misuse. Since we do not add the demographic head to the production model, our method reduces the potential for misuse of this information at deployment. Where one of the alternative models is found to have substantially better fairness properties, it could be substituted for the base model; we would advocate for the removal of the demographic prediction head after training in this circumstance, so as to avoid any potential for misuse.

Our proposed mitigation approaches (subsampling[29,30], gradient reversal) eliminate correlations between sensitive attributes and outcomes, or mitigate their effect on model training. This may seem counterintuitive, particularly where sensitive attributes are thought to be causal drivers of disease. Nevertheless, our framework allows practitioners to identify when such mitigation is desirable by analyzing consequences on the trade-off between model performance and fairness. We however note that the selection of a specific model should be informed by domain knowledge, and multiple other considerations (non-exhaustive list) such as utility, usage, potential distribution shifts[31] and downstream societal factors.

Shortcut learning is usually assumed to involve the encoding of, and reliance upon, spurious or non-causal correlations [10]. As we show, this need not be the case. Shortcut learning may also occur in cases where a sensitive attribute is strongly linked to the clinical task via known, biologically plausible mechanisms. Models may use attribute-clinical correlations appropriately, but these correlations do not generalize to deployment environments. It is also possible that models over-weight the importance of the attribute at the expense of clinical evidence that is more directly predictive, resulting in stereotyping. We chose to focus our study on age for two reasons. Firstly, age is known to be strongly linked to disease risk across a variety of conditions. Secondly, age is grounded as an objective attribute, rather than being socially constructed[29]. These two considerations suggest that age information may be useful in a disease prediction task, making disentangling shortcut learning from appropriate use of input features a more difficult task. For attributes not known to be linked to disease risk, or where attributes are considered to be social constructs, any use of the attribute in the prediction task may be likely to represent a shortcut. In such cases, it may be more appropriate to select demographic parity (independence) as a fairness criterion.

Where shortcutting does occur, there are multiple approaches that can be used to mitigate this particular effect. In our work, we find that gradient reversal can ameliorate, but not fully obviate, the effect of a biased dataset. Balancing the dataset may be an effective strategy, where feasible; we find that balancing the data leads to fair model performance at no cost to overall performance for CXR (see Supplementary Figure 5). Other approaches [24,32,33] could also be considered, although recent work suggests that many strategies improve fairness only by degrading the performance of the model [34]. Regardless of mechanism, ensuring models are as fair as possible remains a vitally important and unsolved challenge for machine learning.

When we applied our framework to a dermatology application, we did not identify a clear pattern of shortcut learning, despite unequal model performance by age. This demonstrates a case in which model performance is not fair; the label is strongly correlated with the sensitive attribute, and the sensitive attribute is encoded by the model. However, the encoding does not appear to be the (main) source of the unfair performance. In this case, unfairness might be caused by other factors, such as:

- Different presentations of the same condition. For instance, the typical pattern of hair loss for females with androgenetic alopecia differs to that of males [35]. Where presentations differ, unfairness can occur due to an inadequate sample size for specific subgroups, or if the appearance of the condition is more "difficult" to identify for some groups. Potential solutions include obtaining more examples of these presentations, upweighting losses for difficult examples, or approaches such as focal loss [36]. There is also evidence that longer training times encourage the learning of more difficult examples [37,38].
- Differences in the quality of the label, or the use of proxy labels to approximate underlying disease. This is extensively discussed in [3,39], with potential mitigations.
- Differences in the quality or missingness patterns of the data. Multiple causes of unfairness in this regard are described in [1].

Lastly, shortcut learning, where present, does not guarantee the absence of other sources of unfair performance.

In our study we focused on disease presence as a binary outcome. It is possible that disease severity or subtype distribution may also be correlated with sensitive attributes, resulting in other forms of shortcut learning. Further work is required to extend our framework to account for these considerations. Similarly, our work considered age as a single attribute of interest. In principle, this method may be readily applicable to an intersectional analysis [40], although practically there may be challenges around model convergence. In addition, ShorT relies on the availability of demographic data and future research should be performed in cases where demographic data is unobserved[41].

Finally, algorithmic fairness is a set of mathematical formulations, and model behavior should be considered in the broader context of health equity, the entire clinical system and its interaction with society, rather than just focussing on model behavior given a defined dataset[42]. In this broader context, it would be informative to compare human to AI performance and fairness; and to consider modeling the therapeutic or clinical consequences of unfairness in diagnostic predictions (for example with decision curve analysis). We however believe that the identification and mitigation of shortcut learning, as demonstrated by our approach, paves the way for more fair medical AI.

# Methods

All images and metadata were de-identified according to Health Insurance Portability and Accountability Act (HIPAA) Safe Harbor before transfer to study investigators. The protocol was reviewed by Advarra IRB (Columbia, MD), which determined that it was exempt from further review under 45 CFR 46.

To identify shortcut learning, and how it relates to the encoding of sensitive attributes by the model, we define multiple quantities: (1) the encoding of the sensitive attribute, (2) fairness metrics, and (3) shortcut testing (ShorT), i.e. the correlation between the encoding of the sensitive attribute and fairness metrics. We demonstrate our proposed approach using binary prediction tasks for findings in an open-source chest x-ray (CXR) dataset. The approach is then applied to a multiclass diagnosis task in dermatology.

## Datasets, tasks and models

For Chest X Ray (CXR) experiments, models were trained using the NIH Chest X Ray dataset [43] to predict a single binary condition label (Figure 7a Condition Prediction) using a cross entropy loss. We investigated the effusion, atelectasis and abnormal findings. Model performance was estimated by computing the Area Under the Receiver Operator Curve (AUROC). The feature extractor was a ResNet 101x3 architecture initialized from BiT checkpoints [44]. We used an Adam optimizer with a constant learning rate and optimal hyperparameters were determined for each class of model before training (see Supplementary Methods). Each model was trained from five different random seeds, with results presented in terms of average and standard deviation across seeds. The code was written in Python v3.9, using Tensorflow [45] v1.15, pandas v1.1.5[46], numpy 1.23.5[47], scikit-learn 1.0.2[48], matplotlib 3.3.4[49], and statsmodels 0.12.2[50].

## Assessing the encoding of demographic information

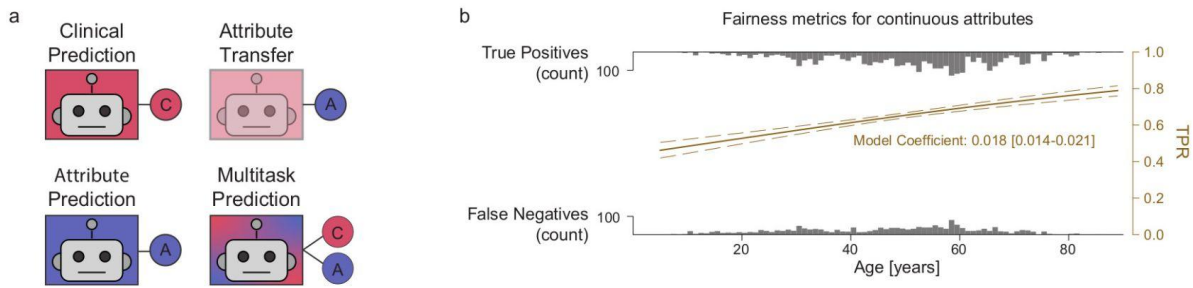

Figure 7: Learning tasks and fairness metrics. (a) Outline of the four learning tasks described in this paper. For each task, a feature extractor backbone (box) is used in conjunction with one or more prediction heads (circles). Clinical and Attribute Prediction tasks use a single head only; Attribute Transfer tasks use a single head with a frozen feature extractor previously trained for a clinical task. Multitask prediction models use both clinical and attribute prediction heads. (b) Logistic Regression (LR) fit for fairness metrics. In this example, a logistic function was fitted to the predictions of the model for examples with the condition. The gray distributions represent counts of true positives (top) and false negatives (bottom) across age (x-axis). The LR model fits the probability of a true positive, as a function of age (TPR, right y-axis in yellow).

Similarly to previous work [13,14], we assessed the encoding of demographic information in the penultimate layer of the model by transferring the condition model to predict age. Once the condition model was trained (Figure 7a, Clinical prediction), we froze all weights in the feature extractor and trained a linear predictor for age (Figure 7a, Attribute Transfer), using a mean squared error regression loss. In order to aid interpretability, the performance of the transfer model was expressed as the Mean Absolute Error (MAE); this value was used as a measure of the age information content of the final layer of the feature extractor, with lower values indicating a more accurate age prediction and hence more age encoding. The results of our experiments were unchanged by using MSE as an evaluation metric. We assess the replicability of this age encoding by training 5 replicates with different random seeds.

To contextualize the obtained MAE, we trained models to directly predict age, without initially training for condition prediction, or freezing the feature extractor (Figure 7a, Demographic Prediction). This provides an empirical lower error bound (LEB) for age for this dataset and model architecture. We estimated an empirical upper error bound (UEB) for age by calculating the error obtained as a result of predicting the mean age of patients in the training set for all examples in the test set (i.e. the baseline performance using the distribution of ages alone, without any image information, in this dataset).

## Assessing fairness for age as a continuous variable

To assess the fairness of a model's predictions according to age, we refer to the definitions in [7] and applied group fairness metrics, quantifying independence, separation and sufficiency. Similarly to [51], we expressed fairness as a function of a continuous attribute variable to avoid the need for quantizing the data [4,12]. This was achieved by fitting a univariate logistic regression (LR) model on age against the outcome of interest (Figure 7b), implemented with Pandas and Scikit-learn[48]. Since we do not wish to assume that the clinical task is independent of age, we focus on separation - the discrepancy of error rates across subgroups, rather than independence - the discrepancy of model predictions across subgroups.

Separation was defined by fitting two LR models to the binarised model predictions, one for patients who do have the condition, and one for patients who do not, equivalent to modeling the effect of age on the True Positive Rate (TPR, sensitivity) and the False Positive Rate (FPR, 1-specificity). The separation value was then calculated as the mean of the absolute values of the two logistic regression coefficients. This resulted in a metric where a score of 0 indicates that there is no monotonic relationship between age and model performance, and higher scores indicate that the TPR and/or FPR vary systematically with age. For small values of separation, our definition approximates the fractional change in performance per year of life - while numerically small, the resultant discrepancy may be large over a clinically relevant age difference. This can be calculated as $e^{s.\Delta a}$, where $s$ is the separation coefficient and $\Delta a$ is the age difference. A separation value of 0.01 will thus correspond to a 10.5% change in model performance per decade; 0.02 will correspond to a 22.1% change per decade.

Our method is agnostic to the choice of fairness definition, and other equivalent formulations could be considered, such as the odds ratio of the logistic regression model, measures computing the maximum gap between subgroups[31] or metrics defined to verify the independence criteria[52]. We however note that it is best to select a metric that will not be dominated by changes in overall performance due to a possible fairness-performance tradeoff (e.g. worst-case group[15]). For completeness, we report multiple fairness metrics in the Extended Data.

## ShorT: testing for shortcut learning

In prior work detecting shortcut learning, the typical assumption is that any effect of the attribute on model output is spurious[21,22]. Methods such as Group-DRO[15] or similar mitigation strategies[53] rely on this assumption and compare model performance across different groups. In the present case, we assume that a difference in performance across groups could be due to a mix of biological and shortcut learning effects (as per our amended definition of shortcut learning). To identify shortcut learning, we hypothesize that if the model is shortcutting, intervening on its encoding of the sensitive attribute should consistently affect fairness metrics beyond the gains in performance. Our goal is hence not to perform a binary (i.e. spurious signal present compared to absent) evaluation, but rather obtaining a continuous modification of the encoding.

Formally, if we assume a feature representation f(X) (here the output of the feature extractor), we want to perform an intervention G such that the relationship between f(X) and the sensitive attribute A is modified between a lower bound (f(X) independent of A) and an upper bound (f(X) strongly related to A). We estimate the efficacy of our intervention with a proxy: the performance of a model predicting A from f(X). Our amended definition of shortcut learning then assesses how the relationship between A and f(X) affects the model's fairness, given a desired minimum performance level.

Based on prior work on adversarial learning[16–19], we used multitask learning to alter the degree to which age is encoded in the penultimate layer of the condition prediction model (Figure 7a, Multitask Prediction). We trained models on both demographic information (here, age) and condition prediction tasks by using a common feature extractor with a separate head for each task. Varying the amount of age information encoded in this model was achieved by scaling the gradient updates from the age prediction head. Positive values of gradient scaling encourage the model to represent age more strongly in the final layer, whereas negative values decrease this representation by gradient reversal. For each model trained (i.e. each combination of gradient scaling and training seed), we assess population variance by bootstrapping test examples 1,000 times for each model trained and report 95% confidence intervals.

We measured the effect of varying the age information encoded in the model by computing model performance, age availability (MAE of the age transfer model, as described above) and fairness metrics for each value of gradient scaling. The presence of shortcut learning is then indicated by a significant relationship between age encoding and fairness metrics. Given our choice of MAE and our formulation of separation (lower is better), we expect a negative correlation (computed via Spearman correlation coefficient) to highlight shortcut learning. For simplicity, we focus on the metrics as evaluated once across the test population for each model, as we observed that population variance was smaller than the variance across models.

## Assessing the efficacy of ShorT to detect shortcut learning

To assess how our method behaves under different bias scenarios, we alter the correlation between the sensitive attribute and the condition label in the CXR training dataset. This was achieved by randomly subsampling the training set [54], with a probability determined as a function of the patient's age and condition label (see Supplementary Methods). After resampling, we obtain two datasets: a biased dataset that introduces approximately a decade of age difference between the positive and negative classes, and a balanced dataset where the distributions of ages across classes are approximately matched. Importantly, whilst this perturbation increased or eliminated the correlation between age and disease, there remain examples of older patients and younger patients with and without the condition, and all retained examples were not modified in any way. Whilst this resampling strategy leads to a slightly lower number of training examples, it has the advantage of maintaining the marginal probability of diseases, and avoids creating synthetic examples which may not be realistic. We expect the biased dataset to lead to strong shortcut learning, while the balanced dataset should lead to no shortcut learning.

In Supplementary Note 1, we also generated simulated data and assessed the sensitivity of the technique when we suspected shortcutting to be absent and when we suspected it to be present.

### ShorT on other datasets

We apply ShorT on two additional datasets: chest x-rays from the public CheXpert database [20] and a proprietary dataset of dermatology images from teledermatology clinics in the United States of America[8]. Model architectures are similar to that described above. Fairness metrics and model hyper-parameters are adapted to the case of a binary attribute (CheXpert) and a multiclass prediction model (dermatology). Please see the Supplementary Methods for details.

## Data availability

The NIH CXR Dataset is provided by the NIH Clinical Center and is available at https://nihcc.app.box.com/v/ChestXray-NIHCC. CheXpert is available at https://stanfordmlgroup.github.io/competitions/chexpert/. Demographic labels are available at https://stanfordaimi.azurewebsites.net/datasets/192ada7c-4d43-466e-b8bb-b81992bb80cf. The de-identified teledermatology data used in this study are not publicly available due to restrictions in the data-sharing agreement. The MNIST[55] data for the simulated experiments is available as a tensorflow dataset at https://www.tensorflow.org/datasets/catalog/mnist and its original version at http://yann.lecun.com/exdb/mnist/.

## Code availability

The deep learning framework (TensorFlow) used in this study is available at https://www.tensorflow.org/. For medical imaging models, the training framework (Estimator) is available at https://www.tensorflow.org/guide/estimators; the deep learning architecture (bit/m-r101x3) is available at: https://tfhub.dev/google/bit/m-r101x3/1. The detailed code to model medical images is proprietary, but pseudocode agnostic to the deep learning framework is in the supplementary material. The code for the simulated experiment using MNIST is available at https://github.com/google-research/google-research/tree/master/shortcut_testing. Scikit-learn is available at https://scikit-learn.org/stable/. We also used Matplotlib (https://matplotlib.org/) for plotting and Pandas for data analysis (https://pandas.pydata.org/).

# Acknowledgements

We thank Alexander D'Amour, Oluwasanmi Koyejo, Stephen Pfohl, Katherine Heller, Patricia MacWilliams, Abhijit Roy, Vivek Natarajan, Charles Lau, Jon Deaton, Dale Webster, William Isaac, Shakir Mohamed, Danielle Belgrave, Greg Corrado and Marian Croak for their contributions to this work. We also thank Lisa Lehmann and Ivor Horn for helpful discussions on the ethical implications of this work. Google funded this work.

## Author contributions

A.B., J.S., N.T., Y.L. and A.K. designed the study, A.B., J.S. and J.F. implemented the method and performed experiments. A.B., J.S., N.T., J.F. and A.K. contributed to the writing of the manuscript.

## Competing Interests Statement

All authors were funded by Google at the time of writing. We declare no competing interests.

# Supplementary Materials

## Supplementary Figures

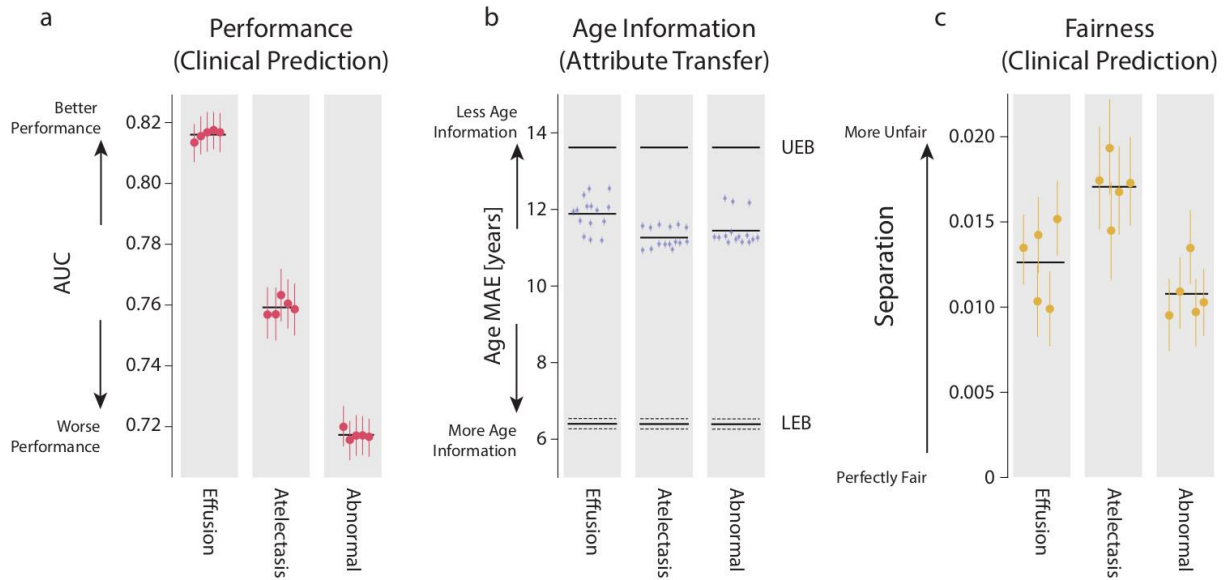

**Supplementary Figure 1: Binary CXR prediction models.** (a) Performance in terms of AUC for each task (Effusion, Atelectasis and Abnormal). (b) Age information encoded in each model in terms of age MAE. UEB: Upper Error Bound, LEB: Lower Error Bound, as determined experimentally. These bounds represent the limits for model age error in this dataset. The LEB is displayed as the mean and standard deviation across 5 technical replicates. (c) Fairness of each model in terms of separation, with 0 meaning a perfectly fair model (a separation value of 0.01 will correspond to a 10.5% change in model performance per decade; 0.02 will correspond to a 22.1% change per decade, see Methods). In all cases, each dot represents a different replicate of the model and error bars represent the population variability (95% bootstrapped confidence intervals, n=17,723 independent samples), with the average metric represented by a horizontal line. Source data are provided as a Source Data file.

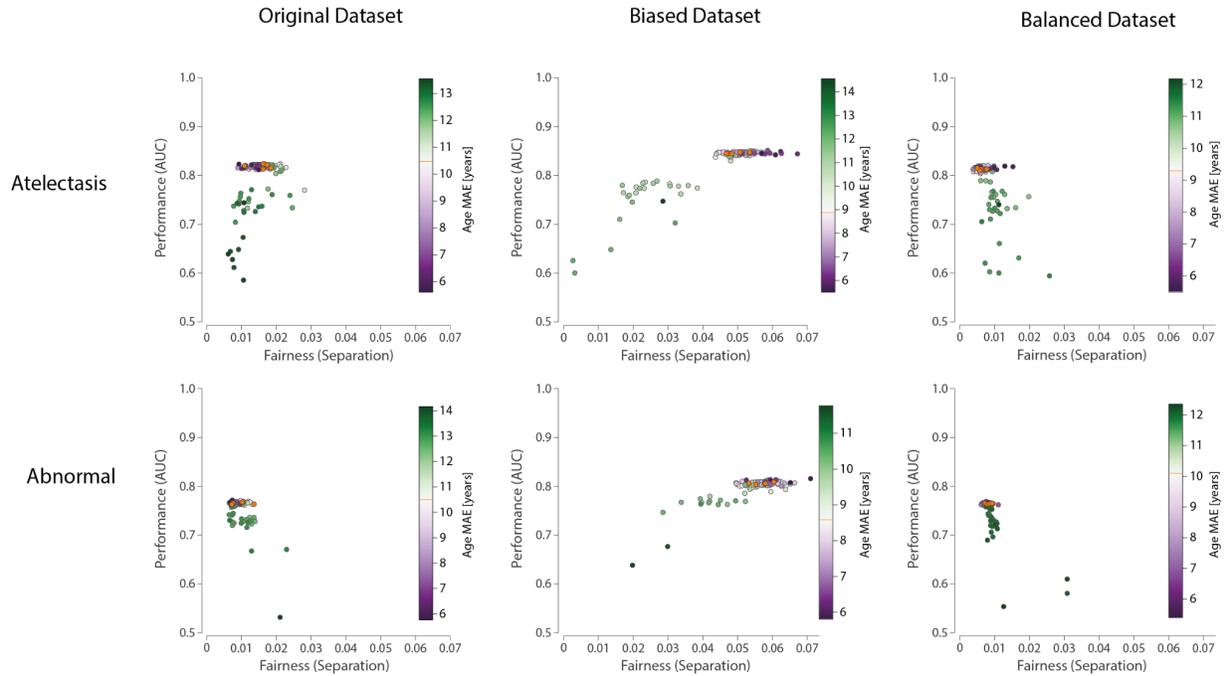

**Supplementary Figure 2: Fairness-Performance results for other CXR labels.** Separation is plotted against AUC, with the age performance of each model represented by the color as in Figure 3. Similar patterns may be observed, whereby inducing a bias in the training dataset results in much more unfair model performance, which can be ameliorated by gradient reversal (center column, green dots), or exacerbated by increasing the age representation (purple dots). In contrast, balancing the training dataset results in baseline models (orange) which are considerably fairer, and gradient reversal results in degraded model performance without further fairness improvement. Source data are provided as a Source Data file.

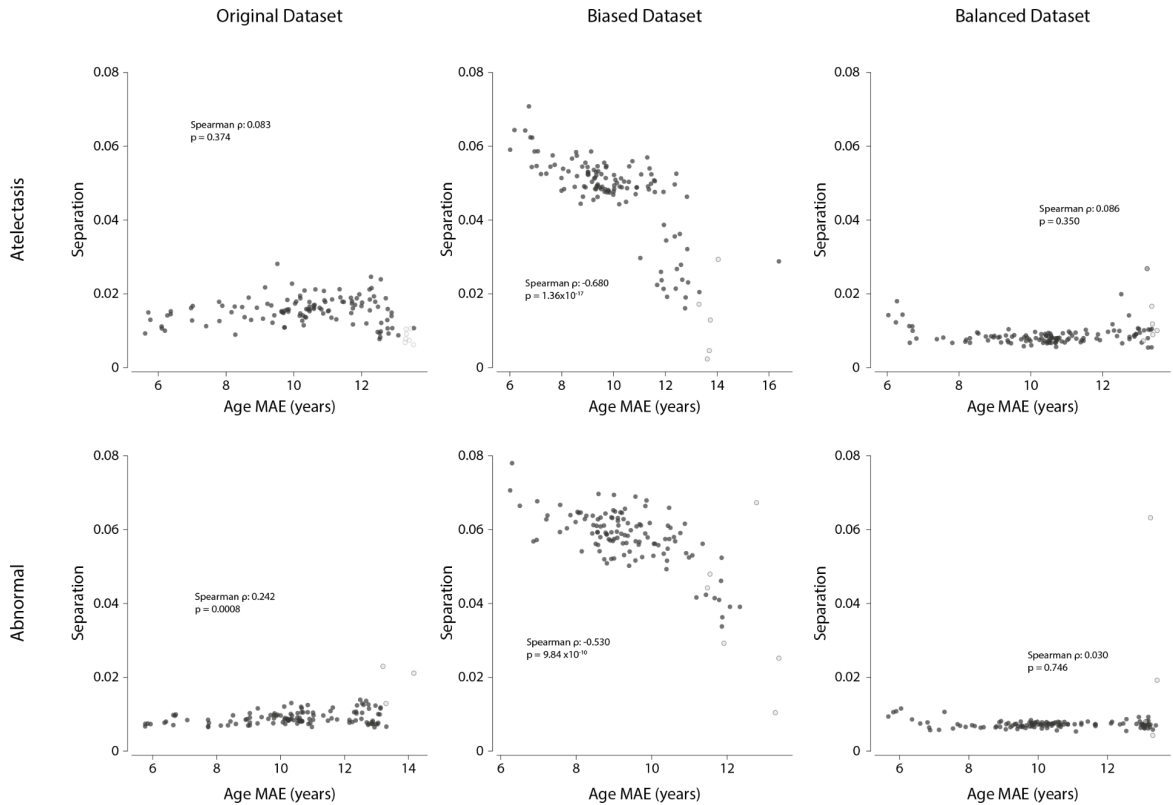

**Supplementary Figure 3: ShorT analysis of original NIH and subsampled datasets for Atelectasis and Abnormal (the complement of the No Finding label).** Biased datasets (middle column) result in significant dependence of fairness on age representation. In contrast, balanced datasets (right column), there is no such dependence. In the original dataset, there is no dependence of fairness on age representation for Atelectasis, however there is a significant positive correlation between fairness and age representation for the No Finding label. This implies that models which represent age more accurately (left) tend to be fairer (closer to 0 on the y axis). This may be explained by an underuse of age information for this particular dataset and task. For all plots, an AUC threshold was set at 0.7, with replicates with an AUC value less than this being excluded from the correlation analysis. We chose 0.7 as the threshold as the performance of baseline models was lower for Atelectasis and Abnormal labels (Supplementary Figure 1). One such replicate is not displayed on the Abnormal, Original Dataset plot, as it had a separation of > 0.1, and lies beyond the limits of the y axis. All tests are two-sided Spearman correlations. Source data are provided as a Source Data file.

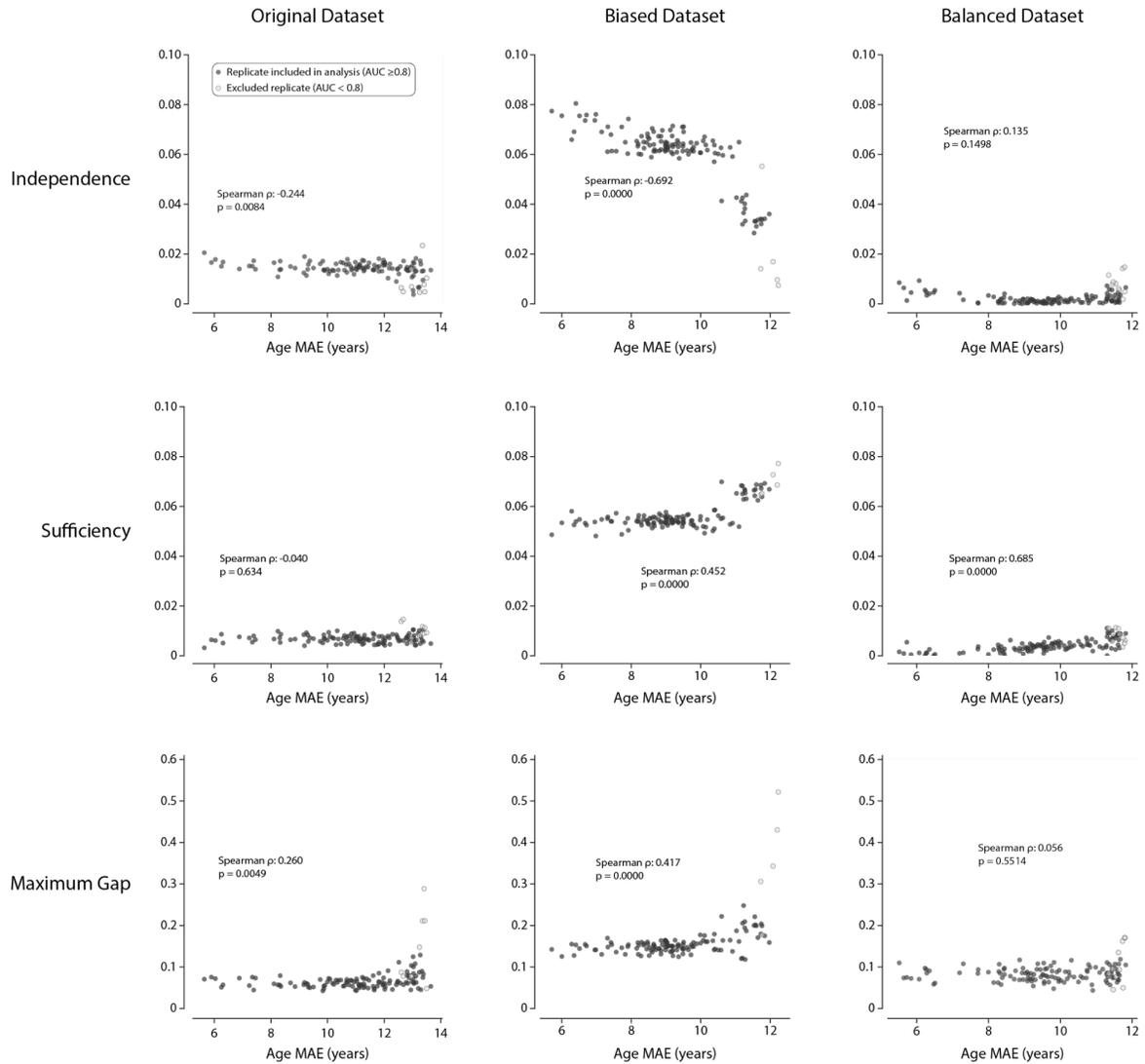

**Supplementary Figure 4: ShorT analysis of original NIH and subsampled datasets for Effusion using different fairness metrics.** (Top row) Independence, as computed by the coefficient of the logistic regression between the model's predictions and age. (Middle row) Sufficiency, as computed by the positive predictive value. (Bottom row) Maximum gap in performance across age subgroups, with age bucketed in [18,30), [30, 45), [45, 65) and [65, 100). For all plots, an AUC threshold was set at 0.8, with replicates with an AUC value less than this being excluded from the correlation analysis. All tests are two-sided Spearman correlations. Source data are provided as a Source Data file.

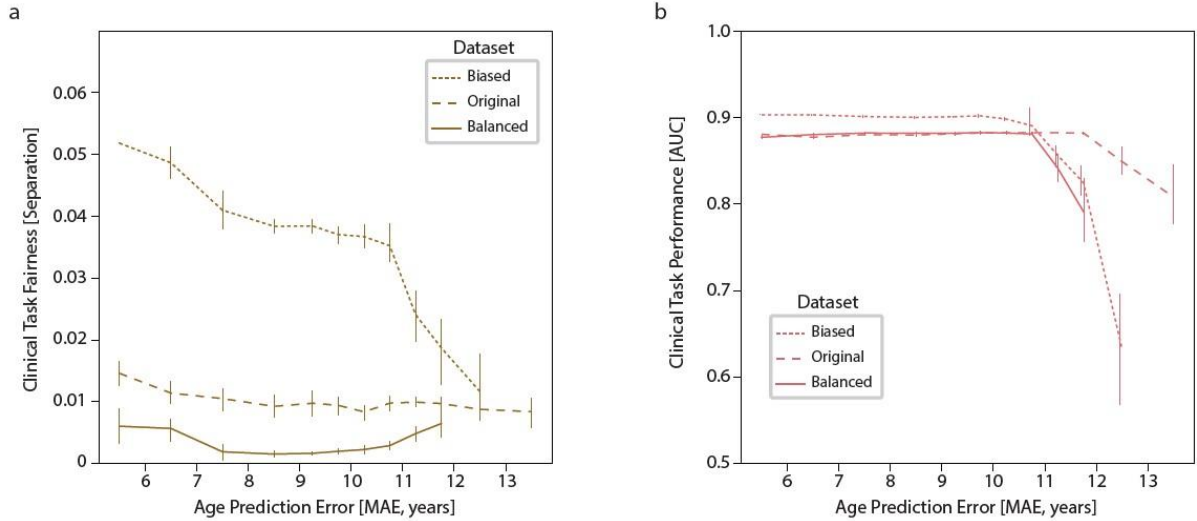

**Supplementary Figure 5: Cross-Dataset Age comparison of fairness and performance at differing levels of age encoding in the Effusion task.** (a) Fairness. Results from Figure 5 are displayed on a single graph, with replicates pooled according to predefined degrees of age encoding. In each degree of age encoding, we define its average (line) and estimate variability across models with bootstrapping (95% confidence intervals, error bars). For a given level of age encoding, models trained on the Balanced (solid line), Original (dashed line), or Biased (dotted line) datasets display vastly different fairness characteristics. (b) Age encoding vs Performance. At the same level of age encoding, performance is very similar for the Balanced and Original datasets, although the performance of the Balanced dataset drops off at higher age prediction errors. The Biased dataset results in a spuriously higher AUC due to cleaner class separation (see Supplementary Figure 6). Source data are provided as a Source Data file (see Figures 2c, 3c and 3f).

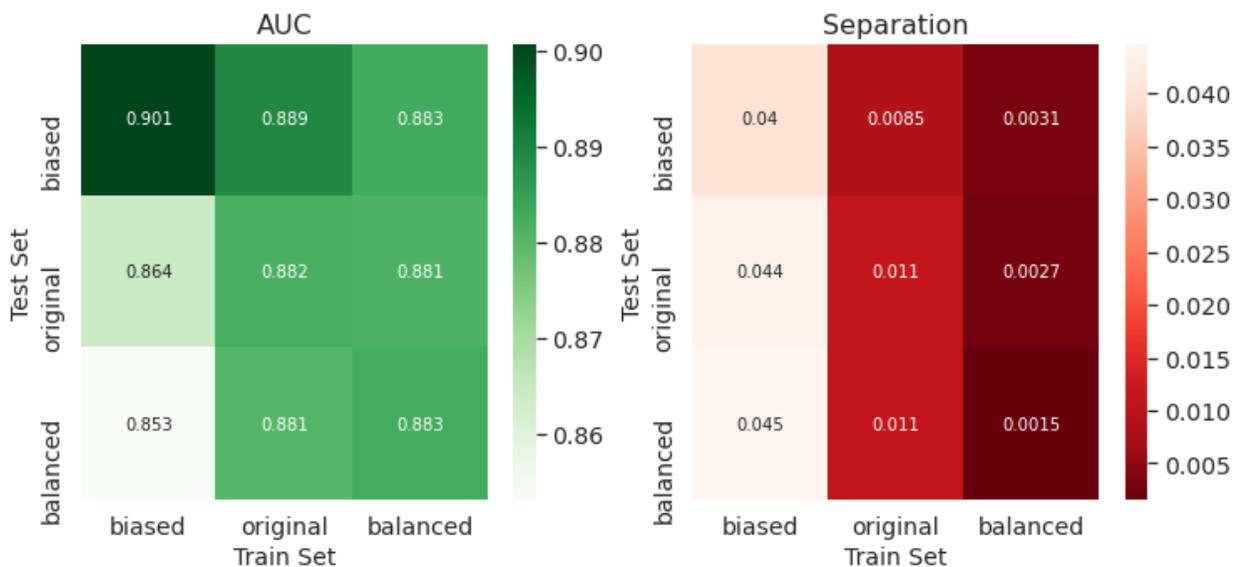

**Supplementary Figure 6: Cross-dataset performance and fairness for the effusion prediction task.** AUC and Separation are shown for baseline models (without an age prediction head) trained on biased, original, and balanced datasets (x axis), tested on all three datasets. In-distribution results are located on the top-left to bottom-right diagonal. Note that the best performance is obtained in models trained on biased datasets, tested in-distribution; however, performance is degraded for out of distribution test sets, due to shortcut learning; this increase in performance is therefore spurious. Models trained on balanced datasets obtain similar performance results to those trained on the original dataset. However, separation is considerably improved in models trained on balanced data.

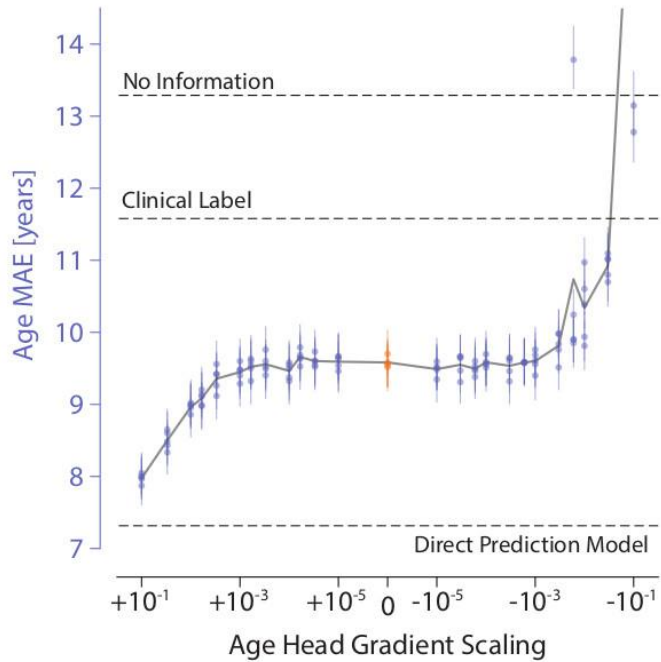

**Supplementary Figure 7: Effect of age head gradient scaling on age representation for the dermatology example in figure 6.** ShorT models covered a range of age prediction errors, although there appeared to be a wider plateau in the middle of the range of age head scaling values, over which age prediction error was quite similar to baseline. This plateau, as well as the wide range between the "Clinical Label" and "No Information" upper error bounds, likely occurs due to the richer soft labels used in this example, as well as due to the stronger dependence between age and condition probability for many (but not all) dermatological problems. Each dot represents a model trained (25 values of gradient scaling times 5 replicates), with error bars denoting 95% confidence intervals from bootstrapping examples (n= 1,925 independent patients) within a model. Source data are provided as a Source Data file.

# Supplementary Methods

## CXR Datasets

We use two CXR datasets, NIH CXR, and CheXpert. The NIH CXR Dataset is provided by the NIH Clinical Center and is available at https://nihcc.app.box.com/v/ChestXray-NIHCC. For experiments, images were first downsized to 448x448 pixels. We select the "Effusion",

"Atelectasis" and "No findings" (which we report as "Abnormal" for semantic consistency) labels provided with data as our binary outcomes, focussing on Effusion.

CheXpert is available at https://stanfordmlgroup.github.io/competitions/chexpert/. Demographic labels are available at https://stanfordaimi.azurewebsites.net/datasets/192ada7c-4d43-466e-b8bb-b81992bb80cf. Following[22], we focus on a binary distinction between Black and White patients, rather than treating race as a multi-class prediction task. Images were downsized to 448x448 pixels, and we select cardiomegaly as a binary outcome for reporting.

The demographic information for the NIH dataset is as follows, broken down by findings:

|  |  |  | Train | Tune |
|---|---|---|---|---|
| No Effusion | Female | Avg age | 47.095458 | 44.460953 |
|  |  | N | 26200 | 16698 |
|  | Male | Avg age | 48.625099 | 44.311849 |
|  |  | N | 32974 | 22931 |
| Effusion | Female | Avg age | 50.663645 | 48.79426 |
|  |  | N | 3199 | 2683 |
|  | Male | Avg age | 51.563483 | 48.064258 |
|  |  | N | 3560 | 3875 |
| No Atelectasis | Female | Avg age | 47.10309 | 44.627401 |
|  |  | N | 26792 | 17335 |
|  | Male | Avg age | 48.547831 | 44.307013 |
|  |  | N | 32751 | 23683 |
| Atelectasis | Female | Avg age | 51.395474 | 48.733138 |
|  |  | N | 2607 | 2046 |
|  | Male | Avg age | 52.059212 | 49.004483 |
|  |  | N | 3783 | 3123 |
| No Finding | Female | Avg age | 46.252016 | 42.902625 |
|  |  | N | 16991 | 9448 |
|  | Male | Avg age | 47.98801 | 43.51865 |
|  |  | N | 21268 | 12654 |
| Finding | Female | Avg age | 49.170374 | 47.113662 |
|  |  | N | 12408 | 9933 |
|  | Male | Avg age | 50.197891 | 46.048544 |
|  |  | N | 15266 | 14152 |

## Model Architectures

For the medical imaging models, we employ convolutional neural networks as image embedding models, followed by multi-layer perceptrons (MLPs) as classification models for both clinical classification and age/race prediction. We use modified ResNet 101x3 architectures[1] pre-trained on the public Imagenet 21k dataset. Model architectures for image embedding and checkpoints are available on tensorflow hub.

In the dermatology task, each clinical case includes 1 to 6 images. We average the embeddings across a clinical case before passing them to the MLP. All clinical classification MLPs have 2 layers, with 512 hidden units and ReLU activation, while all sensitive attribute MLPs have 3 layers, with 512 and 256 hidden units and ReLU activation.

In array pseudocode, the architecture follows:

```
# images: array of size (num_instances, height, width, 3)
# num_classes: 27 if dermatology, else 2
image_embeddings = resnet_101x3(images)
if dermatology:
 image_embeddings = image_embeddings.mean(axis='instance')
clinical_prediction = mlp(
    image_embeddings,
    layers=2,
    units=(512, num_classes),
    activations=('relu', 'softmax')
)
attribute_prediction = mlp(
    reverse_gradient(image_embeddings),
    layers=3,
    units=(512, 256, 1),
    activations=('relu', 'relu', None)
)
if age:
 loss = (cross_entropy(clinical_prediction, labels) + lambda *
        mse(attribute_prediction, sensitive_labels))
elif race:
 loss = (cross_entropy(clinical_prediction, labels) +
        lambda * binary_ce(attribute_prediction, sensitive_labels,
from_logits=True))
```

Where the reverse_gradient is an operation that allows for gradient scaling and lambda is a hyper-parameter (positive or negative) that controls for the strength of the scaling. See the code available at

[https://github.com/google-research/google-research/tree/master/shortcut_testing](https://github.com/google-research/google-research/tree/master/shortcut_testing) for an example implementation of these different operations.

### Hyperparameter Tuning and model selection

All models were tuned for batch size, learning rate, weight decay, and dropout in the penultimate layer before training. The same parameters were applied to models trained on each label in the CXR task.

|  | Batch Size | Learning Rate | Weight Decay | Dropout |
|---|---|---|---|---|
| Age Prediction | 16 | $1 \times 10^{-5}$ | $1 \times 10^{-7}$ | 0 |
| CXR Prediction | 16 | $4 \times 10^{-5}$ | $1 \times 10^{-6}$ | 0.1 |
| Age Transfer | 8 | $3 \times 10^{-3}$ | n/a | 0 |

Models were trained for 17,500 epochs and the model with highest performance on the validation data was selected. The decision threshold for each model was based on the maximum F1-score observed on validation data.

### Multitask Prediction

To adapt a single task prediction model to multitask prediction, we added a demographic (age prediction) head at the final layer of the base model. There are no hidden layers between the feature extractor and the condition output layer. However, the demographic head itself uses two fully connected hidden layers between the gradient reversal layer and the final age output layer, to provide the network with capacity during adversarial training.

Next, in order to approximately balance the losses between the age (mean square error) and condition (cross-entropy) heads, we down-weighted the regression loss by a factor of 100. We then tested further adjustments to this loss weighting using a grid search (in conjunction with a coarse gradient scaling parameter sweep). In our case, we found that simple balancing of losses was sufficient.

Once the loss weighting was established, this was fixed for all further experiments. We then swept over 25 values for scaling of the gradient updates from the demographic head, ranging from -0.1 to +0.1 (spaced exponentially). For each value of gradient scaling, 5 replicates were trained, resulting in 125 models per experiment.

For attribute transfer experiments, the feature extractor was frozen and then a linear demographic prediction head was applied and the model retrained to predict age. Hidden layers were not required in this simpler (single task) prediction setup; we found that the addition of one or two hidden layers made no material difference to our results.

## Subsampling of training data

In order to produce datasets with a shift in the mean age between the ground truth classes, we use a logistic probability function, which defines the probability of an example being retained as a function of the age of the patient:

$$p_{retain} = m \div (1 + e^{-k(a-a_0)})$$

Where $k$ is the slope of the function; $a_0$ is the midpoint of the probability function (the age at which the probability of being retained is 0.5); and $m$ is a scale factor that increases the probability of retaining examples. This defines a probability of retaining a positive example; for negative examples (patients without the condition), we use $1-p_{retain}$

The following parameters were used to generate subsampled training sets. Since the process is stochastic, these were obtained by trial and error.

|  | $k$ |  | $a_0$ | $m$ |
| --- | --- | --- | --- | --- |
|  | Biased | Balanced |  |  |
| Effusion | 0.14 | -0.07 | 50 | 4 |
| Atelectasis | 0.12 | -0.08 | 50 | 4 |
| Abnormal | 0.14 | -0.065 | 50 | 4 |

The training sets generated using these parameters are described below. These perturbed datasets do not precisely match the desired shift in ages due to stochastic errors.

|  | Number of training examples | | | Positive Examples (% of training set) | | | Mean Age of Positive / Negative classes (years) | | | Performance (AUC) Fairness (Separation coefficient) | | |
| --- | --- | --- | --- | --- | --- | --- | --- | --- | --- | --- | --- | --- |
|  | Original | Biased | Balanced | Original | Biased | Balanced | Original | Biased | Balanced | Original | Biased | Balanced |
| Effusion | 65394 | 55634 | 61029 | 6731 (10.3%) | 5612 (10.1%) | 6421 (10.5%) | 51.3 48.2 | 55.8 44.6 | 50.1 50.0 | 0.882 0.011 | **0.901** 0.040 | 0.883 **0.002** |
| Atelectasis | 65394 | 57800 | 59349 | 6354 (9.7%) | 5592 (9.7%) | 5959 (10.0%) | 52.0 48.2 | 55.3 45.3 | 50.6 50.6 | 0.819 0.016 | **0.846** 0.049 | 0.813 **0.006** |
| Abnormal | 65394 | 55128 | 62294 | 27465 (42.0%) | 22299 (40.5%) | 22299 (40.5%) | 50.0 47.5 | 55.1 44.2 | 49.2 49.0 | 0.766 0.010 | **0.806** 0.057 | 0.765 **0.008** |

### Significance testing when comparing ShorT across datasets

Shortcut testing (ShorT) relies on calculating the correlation between the degree of age encoding and fairness metrics. To test that the ShorT statistics differ across datasets, we perform permutation tests of Spearman's rho across different versions of the training dataset. We calculate the true difference in correlation statistics, and compare it to an empirical null distribution of differences. The null distribution is simulated using bootstrapping. We combine the data points from the two groups, shuffle them, and randomly divide them into two groups. To calculate p-values, we compare the true difference to this null distribution.

For CXR, we find that differences are highly significant when comparing the original and biased datasets, and the original and balanced datasets (p = 1e-8; p = 1e-4, respectively), indicating that shortcutting happens significantly more with biased datasets, and significantly less with a balanced dataset.

### Race in cardiomegaly models

To test ShorT in the context of a spurious attribute, we apply it to race in chest x-ray analysis. Following previous work[2], we analyze self-reported race in the CheXpert dataset as a binary task of predicting White and Black self-reported race from chest x-rays. We treat the Uncertain label for Cardiomegaly as negative.

The public validation set available for CheXpert only contains 9 individuals with self-reported Black race. Due to this small sample size, we instead randomly re-split the training data into new training (85%), validation (5%), and testing (10%).

This re-split has the following properties:

|   |   | Tune | | Train | |
|---|---|---|---|---|---|
|   |   | No Cardiomegaly | Cardiomegaly | No Cardiomegaly | Cardiomegaly |
| White | Female | N = 4447<br>Avg Age = 64.9 | N = 491<br>Avg Age = 69.6 | N = 40080<br>Avg Age = 64.9 | N = 4537<br>Avg Age = 70.3 |
|  | Male | N = 6652<br>Avg Age = 62.0 | N = 979<br>Avg Age = 67.1 | N = 60011<br>Avg Age = 62.0 | N = 8427<br>Avg Age = 66.5 |
| Black | Female | N = 480<br>Avg Age = 58.6 | N = 120<br>Avg Age = 61.6 | N = 4140<br>Avg Age = 56.4 | N = 1104<br>Avg Age = 58.6 |
|  | Male | N = 501<br>Avg Age = 57.8 | N = 89<br>Avg Age = 58.2 | N = 4508<br>Avg Age = 54.5 | N = 1028<br>Avg Age = 53.0 |
| Other Race | Female | N = 3116<br>Avg Age = 57.7 | N = 420<br>Avg Age = 61.9 | N = 28065<br>Avg Age = 57.5 | N = 3883<br>Avg Age = 63.2 |
|  | Male | N = 4454<br>Avg Age = 57.2 | N = 609<br>Avg Age = 60.0 | N = 40125<br>Avg Age = 56.1 | N = 5381<br>Avg Age = 59.5 |

We focus on the cardiomegaly prediction, as the cardiomegaly label is imbalanced for race (prevalence for White patients was 11.5%, prevalence for Black patients was 19.8%). Similar to our age models, we train models to directly predict race to estimate the upper bound of performance on race (as per the AUROC on this binary prediction task). We then train models to predict both cardiomegaly and race, while sweeping over the gradient scale for the race prediction head. We set the weight of both heads as equal, as the scale of the loss is the same order of magnitude, and vary the gradient scale between -0.1 and +0.1 to match other experiments in the paper. Using an implementation inspired by Alabdulmohsin et al.[3], we estimate fairness via equalized odds.

### Dermatology Dataset and experiments

For dermatology experiments, models are trained to predict 26 skin conditions with an additional "other" category to capture the long tail of conditions, as a multiclass prediction task, as described in [4]. Our approach differs slightly from previously published results, as we use a more modern architecture (ResNet 101x3 rather than Inception v4), and a slightly smaller training dataset. The commercial dataset used consists of teledermatology images with associated diagnoses obtained by labeling by multiple dermatologists. Unfortunately, this dataset is not available for public use.

We assess model performance for a single class by using binarised metrics. For AUC, we use the prediction score of the chosen class. For separation, we define positive predictions to be examples where the top ranking prediction score is for the chosen class. Using top-3 selection (i.e. a positive prediction is any example where the score for the chosen class is in the top-3 scores) did not change our results.

### Dermatology Dataset - Demographics

Since the dermatological dataset is not publicly available, we report here the basic demographics of the training dataset used. This dataset comprises 12,027 cases obtained from teledermatology clinics in California and Hawaii:

| Attribute | | Percentage in training set |
|---|---|---|
| Race | American Indian / Alaska Native | 0.83 |
| | Asian | 11.6 |
| | Black / African American | 5.99 |
| | Hispanic / Latino | 41.9 |

|  | Native Hawaiian / Pacific Islander | 1.52 |
|---|---|---|
|  | White | 35.5 |
|  | Not Specified | 2.71 |
| Gender | Male | 38.1 |
|  | Female | 61.9 |
| Age | 18-19 | 7.33 |
|  | 20-29 | 22.4 |
|  | 30-39 | 19.2 |
|  | 40-49 | 16.8 |
|  | 50-59 | 19.5 |
|  | 60-69 | 11.6 |
|  | 70-79 | 2.27 |
|  | 80-89 | 0.782 |
|  | 90+ | 0.191 |

# Supplementary Note 1

### Simulated data

We generated simulated data to assess the efficacy of ShorT. The data consisted of MNIST images[5] with the labels representing whether the number hand-written in the image was smaller than 5, or 5 and above. To these images, we added a small colored square at a random location. The color of the square (red or green) could be correlated with the label, and here plays the role of the sensitive attribute A. Noise was added to the image and the square as the tasks were straightforward. We hence obtain a data generating process that corresponds to Figure 1(b). As we control the data generating process, we are also able to generate counterfactual samples, i.e. images for which the color of the square has been switched.

We implemented ShorT with Tensorflow[6] v2 and Keras, using as feature extractor a small MLP of 3 dense layers with 10 units each. For gradient reversal, we added one more dense layer of size 2 before the attribute's output layer while the label was directly predicted from the feature extractor. Attribute encoding was assessed as the ROC AUC after training an output layer from a frozen feature extractor. Fairness was computed via equalized odds. Baseline model accuracy was between 0.8 and 0.86.

Further hyper-parameter selection was needed to balance the losses of the target (weight =1) and of the attribute (search between 0.5 and 1.0). The final value was selected as 0.75. We varied the correlation between Y and A such that a label of Y=0 was associated with a red square between 50 and 95% of the time (20 steps), while the label Y=1 was associated with a red square between 50 and 15% of the time (20 steps).

We observed that ShorT produces significant results for high correlations between Y and A (Supplementary Figure 8a). This corresponds to our observations with counterfactuals that, given the simplicity of the task, the model does not "need" to rely on the attribute for predictions if the correlation between A and Y is not high. Focussing on the low correlation setting, we uniformly sampled the correlations between A and Y in the 0.4-0.6 range (n=50) and assessed the number of significant results for ShorT (at a threshold of $p<0.05$, Bonferroni corrected). We note that only 3 instances lead to significant p-values for ShorT (i.e. 3/50=0.06 ≈ 0.05, Supplementary Figure 8b). Finally, we focussed on the high correlation setting and sampled uniformly in the 0.9-0.98 range for label Y=1 and in the 0.15-0.23 range for Y=0. Note that the asymmetry is needed to obtain unfairness based on equalized odds. In this case, we observe that ShorT correctly identifies shortcutting in all instances (Supplementary Figure S8, 50/50), even after Bonferroni correction for multiple comparisons (50/50).

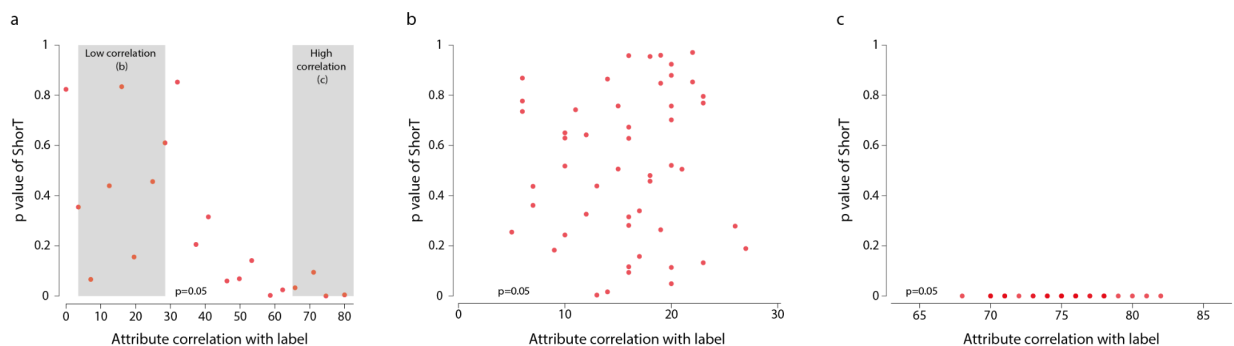

**Figure S8: ShorT on simulated data.** (a) Increasing the correlation between label and color in a consistent but asymmetric fashion leads to significant shortcutting for high values of the correlation. Each dot represents the p-value of ShorT computed based on a different combination of correlations. We focus on two areas (shaded on the plot): a low correlation setting (detailed in (b)) to assess type I error and a high correlation setting (detailed in (c)) to assess type II error. (b) For low values of the correlation and a small asymmetry (x-axis), we obtain a uniform distribution of ShorT p-values. (c) p-values are consistently lower than $p<0.05$ when the asymmetry is high and the correlation between A and Y is large. All tests are two-sided Spearman correlations, with p-values corrected for multiple comparisons using Bonferroni correction.

## Supplementary Discussion

In our analysis, we have chosen to preserve age as a continuous variable, using logistic regression analysis to characterize the fairness properties of the model. This avoids the need for arbitrary quantization of the data. However, it does assume that discrepancies, where observed, will be monotonic - with weaker performance for either older or younger patients. In cases where we may expect bimodal or more complex distributions of fairness properties it might be more judicious to examine the model outputs rather than rely on particular formulations of fairness metrics. Distribution-free approaches [7–9], may be considered if no particular form of association can be expected, although these will in general be more limited in power and interpretability. Secondly, the use of a LR model requires a binarised outcome per example, and would be unsuitable for metrics such as prediction scores (continuous) or AUC (requires a set of observations). Alternative methods [10,11] may overcome some of these limitations, at the expense of interpretability. However, our framework does not require the use of a continuous attribute, and may be applied to binary or discrete variables, by substituting the model-based fairness metrics for conventional definitions.

## Supplementary References